\documentclass{Interspeech2024}

\interspeechcameraready

\title{TOGGL: Transcribing Overlapping Speech with Staggered Labeling}

\name{Chak-Fai}{Li}
\name{William}{Hartmann}
\name{Matthew}{Snover}
\address{BBN Technologies, Cambridge MA, USA}
\email{\{chak.fai.li, william.hartmann, matt.snover\}@rtx.com}


\begin{document}

\maketitle

\begin{abstract}
    
Transcribing the speech of multiple overlapping speakers typically requires separating the audio into multiple streams and recognizing each one independently.
More recent work jointly separates and transcribes, but requires a separate decoding component for each speaker.
We propose the TOGGL model to simultaneously transcribe the speech of multiple speakers.
The TOGGL model uses special output tokens to attribute the speech to each speaker with only a single decoder.
Our approach generalizes beyond two speakers, even when trained only on two-speaker data.
We demonstrate superior performance compared to competing approaches on a conversational speech dataset.
Our approach also improves performance on single-speaker audio.
\end{abstract}

\section{Introduction}
\label{sec:intro}

A major challenge for automatic speech recognition (ASR) is the presence of multiple overlapping speakers, especially when only one channel is available.
In meeting datasets the percentage of overlapped speech can be as high as 13\% \cite{ccetin2006analysis}.
When moving to conversational speech, especially when multiple independent conversations are occurring, the percentage of overlap can be even greater \cite{barker2018fifth}.
Even if the speech does not overlap, standard ASR systems will simply mix the speech from the multiple speakers into a single transcription.
A diarization step is required to first separate the speech based on the individual speakers.
When two or more speakers are speaking simultaneously, then even traditional diarization is not enough as the overlap must still be detected and mitigated \cite{bullock2020overlap}.

Separation and recognition of multi-party speech, also known as the cocktail party problem \cite{cherry1957human}, has a long history.
A number of approaches have been proposed over the years.
One approach is a speaker separation-based approach.
Prior to performing ASR, the signal is separated into two or more signals where each signal contains speech from a single speaker \cite{menne19analysis, chen2021continuous}.
Speaker separation performance has greatly improved in the last few years, even in the case of a single microphone channel \cite{luo2019conv, subakan2021attention, wang2023tf}.
Once the speech has been separated, ASR is performed on each of the separated signals.
While these approaches can be successful, they potentially introduce a delay as each separated signal must be processed separately after separation.
There is also the potential for a cascade of errors as the ASR system is unable to recover from errors made during separation, though joint training can help mitigate this issue \cite{von2020end}.
A similar approach is target-speaker ASR \cite{moriya2023streaming, zhang2023conformer}.
If the individual speakers in the mixture can be identified, then a representation for each speaker (either an embedding or speech example) is provided to the ASR system during decoding.
Inference still needs to be performed once per speaker.

An alternative approach is to either jointly separate and recognize, or to not perform any explicit separation.
If the number of speakers is known \textit{a priori}, or we at least know the maximum number of possible speakers, then we can train a system with one output layer or decoder per speaker \cite{yu2017recognizing, tripathi2020end, chang2019end, lu2021streaming}.
Each output layer only transcribes the speech from one speaker in the mixture.
While each output layer can potentially process speech in parallel, it does increase the complexity of the model and the total number of parameters.
Chang et al. \cite{chang2019end} use a single decoder, but two independent attention layers, one for each speaker.
In order to address the ordering of the speakers, a permutation invariant training (PIT) objective is used.
Lu et al. \cite{lu2021streaming} explicitly build separation into the network through the use of a masked-based encoder that separates the two speakers.
Each encoded representation is processed by a separate RNN-T decoder.

Another approach is to serialize the output.
The model uses a single output layer or decoder to recognize the speech from all speakers.
Special tokens or tags must be used to indicate which speaker each output token is associated with.
One question is the level of serialization.
In \cite{kanda2020serialized}, they serialize at the utterance level; all words from a single speaker are recognized before moving to the next speaker.
This presents a challenge for the model as it must remember the output and speakers as it backtracks to the beginning of the audio for each speaker.
In \cite{kanda2022streaming}, serialization is performed at the level of tokens.
The next token for each speaker is transcribed iteratively, with a special token that indicates a shift between speakers.
The difference with \cite{kanda2020serialized} is the shift can occur any time within an utterance and it can occur multiple times during decoding.
The approach is also extended to a larger number of speakers by using special tokens to identify the output with one of $n$ predefined speakers.
More recent work has also integrated speaker attribution through token-level speaker embeddings \cite{kanda2022streaming-speaker}.
The serialized approach has also been extended to the problem of joint transcription and translation of speech \cite{papi2023token}.

We propose a model for \textbf{T}ranscribing \textbf{O}verlapping speech with sta\textbf{GG}ered \textbf{Labeling}, or \textbf{TOGGL} for short.
Our TOGGL model builds upon the serialization approach, but with a critical design difference.
We introduce special tokens that indicate switching either to the next or previous speaker in the output.
The switching tokens allow the model to generalize to a potentially unlimited number of speakers.
When only a single speaker is present in the signal, our model also acts like a standard ASR model.
We combine the TOGGL model with a mixture-aware self-supervised learning approach \cite{fazel2023cocktail} for pretraining.
As the pretraining significantly improves performance, we include it for all comparisons against competing approaches.
The major contributions of our work are as follows:

\begin{itemize}
    \item The TOGGL model simultaneously recognizes the speech from multiple overlapping speakers.
    \item Demonstration of performance on difficult conversational speech settings.
    \item Training on overlapping speech improves performance on single-speaker data.
\end{itemize}

\section{Approach}

\subsection{Speaker Switching}

\begin{figure}[h]
    \fbox{\includegraphics[width=0.42\textwidth]{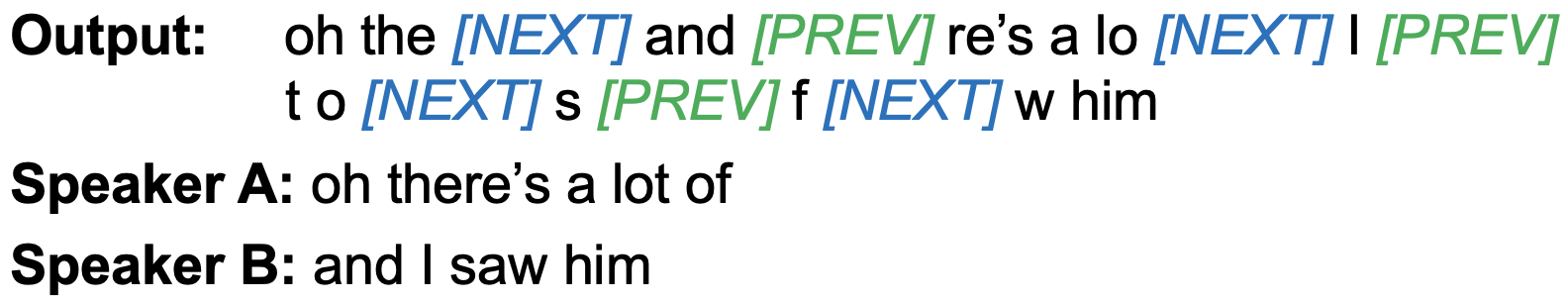}}
    \caption{\label{fig:toggl-decode-example} {\it Example decoding output including the special [NEXT] and [PREV] tokens. Given this output, it can easily be separated into utterances for the two speakers.}}
\end{figure}

The TOGGL approach is essentially a serialization-based approach.
As discussed in the Section \ref{sec:intro}, a major design choice is the level of serialization.
Our preliminary investigations have shown that moving the level of serialization from utterances to words to tokens progressively improves overall performance, so we focus on token-level serialization in this work.
In order for the TOGGL model to jointly recognize the speech from multiple speakers, we introduce two new output tokens: \textit{[NEXT]} and \textit{[PREV]}, which we refer to as TOGGL tokens.
During recognition, all regular output tokens are initially attributed to the first speaker.
When the [NEXT] token appears, the subsequent tokens are attributed to the second speaker.
The model can move to a third speaker by generating another [NEXT] token, or return to the first speaker by generating the [PREV] token.
An example of this process is shown in Figure \ref{fig:toggl-decode-example}.
Skipping between speakers is accomplished by generating multiple [PREV] or [NEXT] tokens sequentially.

\subsection{Self-Supervised Pretraining}

For pretraining, we adopt the the same approach as Cocktail HuBERT (C-HuBERT) \cite{fazel2023cocktail}.
The C-HuBERT approach is an extension of the HuBERT \cite{hsu2020hubert} pretraining approach for mixtures of audio.
Given a corpus of audio consisting of non-overlapping speech, extract a feature vector for each frame.
As in \cite{li2022combining}, we find using the encoder from a supervised model is a better starting point for pretraining compared to the MFCC features used in the original HuBERT study.
Once we have the features for the associated audio, we perform k-means clustering (with 5000 clusters) to generate unsupervised targets.
These initial steps are all done with a single speaker corpus.

Once we have unsupervised targets associated with each frame of audio, we mix the audio to create overlapping speech.
For any frame of audio, we have two or more unsupervised targets, depending on the number of sources mixed together.
During pretraining, we use $K$ projection heads---where $K$ is the maximum number of speakers in a mixture---as opposed to the single projection head used in HuBERT.
The multiple projection heads allow the model to predict each of the unsupervised targets associated with the mixture.
In the original C-HuBERT study, they use Permutation Invariant Training (PIT) \cite{yu2017permutation} because the order of the predicted targets is not important.
In our implementation, we found PIT to have a negative impact on the final performance of the model, so we have omitted it in this study.

\subsection{Supervised Fine-Tuning}
\label{sec:sft}

Once the pretraining phase is completed, we have a pretrained speech encoder.
The next step is to finetune the model for ASR.
We pair the pretrained encoder with a randomly initialized autoregressive decoder and jointly train with CTC \cite{kim2017joint}.
The critical decision is how to represent the labels for training the model.

\begin{figure}[h]
    \fbox{\includegraphics[width=0.42\textwidth]{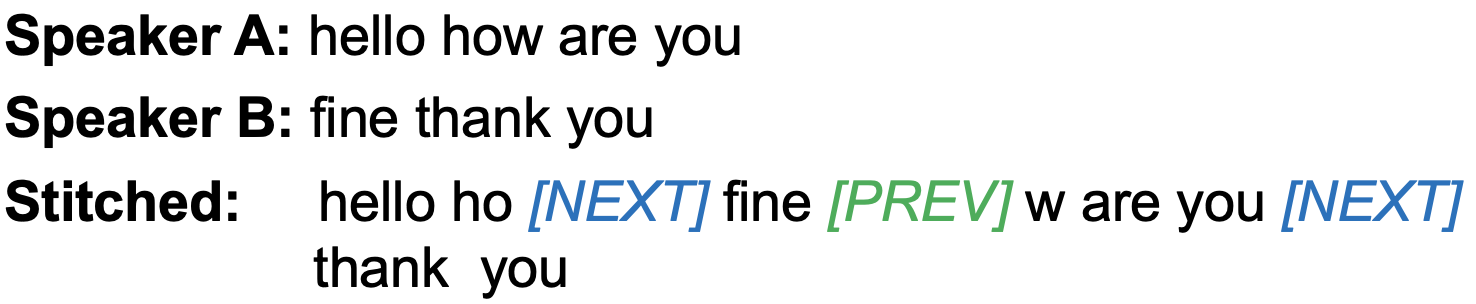}}
    \caption{\label{fig:toggl-train-example} {\it Example of two training utterances being stitched together into a single transcription using the special [NEXT] and [PREV] tokens.}}
\end{figure}

Based on preliminary experiments, we found the best performance when the TOGGL model was allowed to switch between speakers at the token level when comparing to sub-word or utterance level switching, in low to medium resource condition that we are focusing on.
For any given mixture, the transcripts of the individual sources must be merged and the special [NEXT] and [PREV] tokens must be introduced.
We force-align the individual sources to generate time-aligned transcripts at the character level.
Based on these timings, the transcripts are interleaved and the appropriate tokens for switching between speakers is added.
An example is shown in Figure \ref{fig:toggl-train-example}.

As opposed to pretraining, we did find improvement from PIT.
However, we still used a consistant heuristic to specify the speaker order in the transcripts.
The heuristic we used is for the speaker order match the order of the first output character in the time-aligned transcripts.
Whichever speaker spoke first was considered the first speaker, regardless of the duration.

Our approach does introduce issues with regards to the CTC objective.
Because CTC generates one token per frame, it cannot handle the case where the number of output tokens is greater than the number of input frames \cite{graves2006connectionist}.
When we have multiple overlapping speakers, this situation can arise.
We overcome this issue by duplicating the encoder inputs used by the decoder.
For each embedded representation generated by the encoder, we repeat it $n$ times in interleaved fashion,  based on the average speaking rate and number of speakers in the speech. 
For simplicity, we set $n$ equal to the maximum number of speakers seen in training in our experiments.
To be clear, this modification does not leak information about the number of speakers in a particular utterance to the model.
During inference, even if the utterance contains only a single speaker, we still perform this duplication process.
We also noticed the CTC decoder struggles to predict the correct TOGGL token, so we remove them from the reference target only for the CTC objective function.
The CTC decoder only needs to predict the tokens without the special TOGGL tokens; we rely only on the autoregressive decoder for the correct attribution of speakers.
We believe the issues with CTC could also be overcome by replacing the CTC objective with an RNN-T \cite{graves2012sequence} and we plan to explore this in future work.

\section{Experimental Setup}

\subsection{Data Setup}

The majority of work in overlapped speech recognition evaluates performance on read speech that as been artificially mixed, such as WSJ0-mix \cite{isik2016single}, LibriMix \cite{cosentino2020librimix}, and LibriCSS \cite{chen2020continuous}.
We instead focus on conversational speech and use a subset of the Fisher and Switchboard English data available through the LDC\footnote{https://catalog.ldc.upenn.edu/\{LDC2004S13, LDC2004T19\}}.
The full set of Fisher and Switchboard consists of 1700 hours and 300 hours of CTS data, respectively.
We select a 500 hour split to use in our training.
An additional 4 hours is held out for testing.
We believe this setup is both a more challenging and more realistic condition compared to read speech.


While the corpora contain conversations between multiple speakers, each channel only consists of a single speaker.
We artificially mix the speech to create multi-speaker training and test sets.
For each utterance, we mix it with randomly selected utterances from one to two additional speakers, with a random offset and energy modification.
The random offset of the $n$-th utterance in each mixture is drawn uniformly between 0 and 90\% of the length of ($n-1$)-th utterance to form a mixture with 10 to 100\% of its length containing overlapping speech.
The offset of the first utterance is always 0.
The random energy modification of the $n$-th utterance in each mixture is drawn uniformly between -3dB and 3dB of the ($n$-1)-th utterance, and energy of the speech mixture will be normalized to the same as the energy of the initial utterance. 

\subsection{ASR Models}

All of our ASR models are conformer \cite{gulati2020conformer}-based encoder-decoder models trained using ESPNet \cite{watanabe2018espnet}.
We adopt a similar configuration as used in \cite{guo2021recent}.
The encoder consists of 12 layers with four attention heads per layer and an embedding dimension of 384.
The dimension of the feed forward component in each conformer layer is 2048.
The decoder consists of 6 layers with an identical configuration.

We compare the performance of the TOGGL model against several alternative approaches.
Each approach shares the same model structure described above, but differs in either how it is trained or in the number of decoders.
While the competing approaches are similar to previous work discussed in Section \ref{sec:intro}, we use our own implementation.
Each also has the advantage of the same pretraining approach, including the Baseline model; even though the Baseline is only finetuned to transcribe a single speaker, it still starts from a model pretrained on overlapping speech with the C-HuBERT approach.
The specific details of each model are described below.

\textbf{Baseline:} Standard encoder-decoder model finetuned only on non-overlapping speech.

\textbf{Dual-Decoder:} A model with two independent autoregressive decoders, supporting up to two overlapping speakers.

\textbf{t-SOT:} Trained with a special token to allow the model to switch between two speakers at the token level during decoding.

\textbf{TOGGL (2-mix):} Our proposed model with the [NEXT] and [PREV] tokens trained only on up to two overlapping speakers.

\textbf{TOGGL (3-mix):} Same as the (2-mix) version, but trained on up to three overlapping speakers (including during pretraining).

\section{Results}

\subsection{Overall Performance}

\begin{table}
    \caption{\label{tab:english} {\it WER(\%) for each of the models considered. The $n$-mix columns refer to the number of overlapping speakers in the test set, where 1-mix refers to the standard single-speaker test set. The starred (*) entries are an oracle result where only the two speakers that minimize WER are considered.}}
    \centering
    \begin{tabular}{lcccc}
    \toprule
        Model Type & 1-mix & 2-mix & 3-mix & 4-mix \\
        \midrule
        Baseline & 18.6 & --- & --- & --- \\
        Dual-Decoder & 12.0 & 18.5 & 53.7* & 86.5* \\
        t-SOT & 11.8 & 19.2 & 51.4* & 79.0* \\
        TOGGL (2-mix) & 11.2 & 19.1 & 42.3 & 56.9 \\
        TOGGL (3-mix) & 11.4 & 17.7 & 30.7 & 40.6 \\
    \bottomrule
    \end{tabular}
\end{table}

Performance for the various approaches are shown in Table \ref{tab:english}.
In all cases the models are trained on the same 500 hours of English CTS data.
For the models specifically designed to handle overlapping speech (i.e. Dual-Decoder, t-SOT, and TOGGL), we create additional training data by mixing audio from the same pool of 500 hours of data.
For the TOGGL (3-mix) case, the model is also trained on a mixture of three speakers from the same pool of data.
While some of the models (i.e. t-SOT and Dual-Decoder) do not have the capacity to generate output for more than two speakers, we still report results for the 3-mix and 4-mix case.
In those cases, for each utterance, we select the two speakers that minimize WER and ignore additional speakers for the WER calculation.
This represents an optimistic score that does not penalize the model for its inability to produce output for more than two speakers.

The Baseline model achieves a WER of 18.6\% on the 1-mix test set.
All of the models designed to handle overlapping speech outperform the model the baseline on the 1-mix test set.
This is not due to pretraining as the Baseline model starts from the same pretrained model as the others.
The improvement comes solely from the finetuning.
We believe the improvement results from the model seeing different views of the training data, similar to other types of data augmentation.

When considering the 2-mix test set, we see all models perform similarly.
While there is a degradation overall compared to the 1-mix test set, the models do demonstrate the ability to separately recognize the speech from two overlapping speakers.
It is interesting to directly compare the t-SOT and TOGGL (2-mix) models.
The only substantial difference between the two models is that the TOGGL model requires two special tokens and the t-SOT model requires one.
Despite this difference, performance is nearly identical.
However, the TOGGL (3-mix) model provides the best performance overall.

The true benefit of the TOGGL models is only revealed when considering the 3-mix and 4-mix test sets.
Even when the model has not been trained to on these conditions, it is able to generalize.
The WER improvement compared to the t-SOT and Dual-Decoder models is substantial.
This is even more impressive when we consider the WER calculation for those models only considers a maximum of two speakers, while we score the TOGGL models against all speakers.
There is also a large improvement between the 2-mix and 3-mix varieties of the TOGGL model.
While the TOGGL model has the ability to generalize to a larger number of speakers than seen in training, it does benefit from seeing more overlapping speakers in training.
We have also replicated these results on other languages, but do not include those results due to space and time constraints.

\subsection{Overlap Percentage Breakdown}

\begin{table}
    \caption{\label{tab:overlap} {\it WER(\%) broken down by overlap percentage based on the 2-mix dataset. Each column a different range of overlapping speech. For performance on non-overlapping speech or the overall average, see the 1-mix and 2-mix columns in Table \ref{tab:english}, respectively. }}
    \centering
    \begin{tabular}{lccccc}
    \toprule
     & \multicolumn{5}{c}{Overlap Percentage (\%)} \\
    \toprule
        Model Type & 0-20 & 20-40 & 40-60 & 60-80 & 80-100 \\
        \midrule
        Dual-Decoder & 12.7 & 16.3 & 19.9 & 23.1 & 34.4 \\
        t-SOT & 12.3 & 16.2 & 20.9 & 25.7 & 34.5 \\
        TOGGL (2-mix) & 12.1 & 16.4 & 20.5 & 25.7 & 34.3 \\
        TOGGL (3-mix) & 12.5 & 15.7 & 19.3 & 22.9 & 28.6 \\
    \bottomrule
    \end{tabular}
\end{table}

A more detailed breakdown of the performance on overlapping speech can be seen in Table \ref{tab:overlap}.
We group each utterance in the 2-mix test set according to the amount of overlap.
Each column in the table represents a different grouping from 0-20\% to 80-100\%.
Performance across the models in each of the overlap groupings is similar.
The only significant difference is with the TOGGL (3-mix) model in the higher overlap conditions.
This shows that the improvement from that model comes from its ability to perform better in higher overlap conditions.




\begin{table}
    \caption{\label{tab:ablation} {\it WER(\%) for best TOGGL model where each row represents a single change in either the model or the training process. In all cases the models are tested on the 1-mix and 2-mix test sets. }}
    \centering
    \begin{tabular}{lcc}
    \toprule
        Model Description & 1-mix & 2-mix  \\
        \midrule
        TOGGL (3-mix) & 11.4 & 17.7 \\
        \hspace{2mm} + PIT (pre-train) & 14.1 & 25.7 \\ 
        \hspace{2mm} - PIT (fine-tune) & 11.9 & 20.4 \\
        \midrule
        \hspace{4mm} - CTC Enhancement & 15.7 & 29.5 \\
        \hspace{4mm} - CTC & 17.3 & 23.0 \\
        \midrule
        \hspace{6mm} - 3-mix data & 23.8 & 30.7 \\
    \bottomrule
    \end{tabular}
\end{table}

\subsection{Ablations}

The final TOGGL model consists of a number of design decisions.
Table \ref{tab:ablation} reports a series of ablation experiments that illustrates the impact of each individual design decision.
Several of the design decisions have a major impact on overall performance.
The first two rows show that introducing PIT during pretraining significantly hurts performance.
Similarly, removing PIT during finetuning also hurts performance, but not to the same extent.
Starting from the model without PIT during finetuning, we test two modificaiton to CTC.
Removing our proposed CTC enhancement (c.f. Section \ref{sec:sft} brings a large increase in WER.
Removing CTC altogether also hurts performance.
This shows that the presence of the CTC objective hurt performance overall without the proposed changes.
Finally, building on top of the model without CTC and PIT in fine-tuning, we remove the 3-mix data during fine-tuning---3-mix data is still present during pretraining.
This produces a further large degredation in performance.
The ablation experiments demonstrate each of the design decisions involved in training the TOGGL model have a large impact on overall performance.

\section{Conclusions}

We have introduced a novel approach to simultaneously transcribing multiple overlapping speakers in a single-microphone setting.
Our proposed TOGGL model, along with several alternative approaches, is tested on a conversational speech scenarios with up to four overlapping speakers.
We demonstrate that our TOGGL model outperforms previous approaches, generalizes to conditions it is not trained on, and even improves performance on single-speaker audio.
One limitation of the current approach is the reliance on artificially mixed data in training.
In future work we plan to explore approaches to self-supervised pretraining that move beyond the use of artificially mixed data to data mixtures collected in the wild.


\bibliographystyle{IEEEtran}
\bibliography{mybib}

\end{document}